\pdfoutput=1
\documentclass[pdflatex,times,twocolumn,final,authoryear]{elsarticle}

\usepackage{prletters}
\usepackage{caption}
\usepackage{latexsym}
\usepackage{multirow}
\usepackage{graphicx}
\usepackage{array}
\usepackage{subcaption}
\newcolumntype{C}[1]{>{\centering\let\newline\\\arraybackslash\hspace{0pt}}m{#1}}
\newcolumntype{L}[1]{>{\raggedright\let\newline\\\arraybackslash\hspace{0pt}}m{#1}}
\usepackage{url}
\usepackage{xcolor}
\definecolor{newcolor}{rgb}{.8,.349,.1}
\usepackage{amsmath}
\usepackage{amssymb}
\usepackage{bbold}
\usepackage{bbm}

\journal{Pattern Recognition Letters}

\begin{document}

\begin{frontmatter}


\title{Appearance-based indoor localization: A comparison of patch descriptor performance}

\author[1]{Jose Rivera-Rubio\corref{cor1}} 
\cortext[cor1]{Corresponding author:
  Tel.: +44 (0)20 7594 5463;}
\ead{jose.rivera@imperial.ac.uk}
\author[1]{Ioannis Alexiou}
\author[1]{Anil A. Bharath}

\address[1]{Imperial College London, South Kensington Campus, London SW7 2AZ, United Kingdom}

\received{21 Feb 2014}
\finalform{2 March 2015}
\accepted{9 March 2015}
\availableonline{11 March 2015}

\begin{abstract}
Vision is one of the most important of the senses, and humans use it extensively during navigation.  We evaluated different types of image and video frame descriptors that could be used to determine distinctive visual landmarks for localizing a person based on what is seen by a camera that they carry.  To do this, we created a database containing over 3 km of video-sequences with ground-truth in the form of distance travelled along different corridors.   Using this database, the accuracy of localization -- both in terms of knowing which route a user is on -- and in terms of position along a certain route, can be evaluated.  For each type of descriptor, we also tested different techniques to encode visual structure and to search between journeys to estimate a user's position. The techniques include single-frame descriptors, those using sequences of frames, and both colour and achromatic descriptors. We found that single-frame indexing worked better within this particular dataset.  This might be because the motion of the person holding the camera makes the video too dependent on individual steps and motions of one particular journey.  Our results suggest that appearance-based information could be an additional source of navigational data indoors, augmenting that provided by, say, radio signal strength indicators (RSSIs). Such visual information could be collected by crowdsourcing low-resolution video feeds, allowing journeys made by different users to be associated with each other, and location to be inferred without requiring explicit mapping. This offers a complementary approach to methods based on simultaneous localization and mapping (SLAM) algorithms.
\end{abstract}


\end{frontmatter}


\section{Introduction}
\label{sec1}

Satellite-based global positioning systems (GPS) have been available to consumers for many years. When combined with other sensor data, such as terrestrial-based radio signal strength indicators (RSSI), the quality of pedestrian localization within cities, at street level, can be quite reliable.  Recently, interest has been gathering for the development of systems for indoor position sensing: we might consider this to be the next challenge in localization systems \citep{Quigley2010,TUMindoor,Shen,Kadous2013}. Indoor sensing is likely to require additional infrastructure, such as
Bluetooth-based RSSI, or time-of-flight systems. At the time of writing, both of these are reported to be under trial.

Despite this, vision-based navigation systems are under active development.  This might be because such systems do not require special markers to be embedded within the environment.  However, another reason could be that vision provides an immensely rich source of data, from which estimating position is also possible.  For example, in emerging applications \citep{Moller2012} such as mobile domestic robots, merely knowing where a robot is is not enough: a robot often needs some information about its immediate environment in order to take appropriate decisions. This includes navigating around obstacles and identifying important objects (e.g. pets).

Systems and devices that are designed to help humans to navigate would be improved by incorporating vision as one of the sensing modalities. This is particularly true of systems that are designed to help visually impaired people to navigate (assistive navigation systems). However, for wearable or smartphone-based systems, accuracy and power consumption remain two of the challenges to the reliable and continuous use of computer vision techniques.  Visual localization accuracy is affected by several factors, including the techniques used to infer location from visual data. In the case of feature-based SLAM \citep{Davison2003}, for example, a lack of features, or a highly occluded environment, can reduce accuracy.

\cite{Wang2012} have recently suggested an interesting approach to localization based on the principle of identifying landmarks in ambient signals. These ambient signals are acquired from a crowdsourcing-like approach, rather than explicitly mapping out signal strength and WiFi identifiers and appears to offer good performance, with median absolute localization error of less than 1.7 m. Perhaps more importantly, it removes the need to change building infrastructure specifically for localization.  One way to strengthen the landmark approach would be to incorporate {\it visual} cues, automatically mined from the video data.  
Theoretically speaking, such an approach might be limited by i) the quality
 of the image acquisition, which could be affected by blur, poor focusing, 
inadequate resolution or poor lighting; ii) the presence of occlusions to 
otherwise stable visual landmarks; iii) visual ambiguity: the presence of 
visually-similar structures, particularly within man-made environments.
 
We now consider something of an ideal situation in which we can harvest  
visual signatures from several journeys down the same route; the approach starts with the idea of collecting {\em visual paths}, and using the data from these to localize the journeys of users relative to each other, and to start and end points.

\section{Visual Paths}
Consider two users, Juan and Mary, navigating at different times along the same notional path.  By notional path, we refer to a route that has the same start and end points.  An example of indoor  notional paths would be the navigation from one office to another, or from a building entrance to a reception point.  For many buildings, such notional paths might allow different \textit{physical} trajectories which could diverge. For example, users might take either stairs or lifts, creating path splits and merges.  Such complex routes could be broken down into route (or path) segments, and path segments could contribute to more than one complete notional path.

For any notional path or path segment, both humans and autonomous robots would ``experience'' a series of cues that are distinctive when navigating along that path.  In some instances, however, the cues might be ambiguous, just as they might be for radio signal strength indicators, audio cues and other environmental signals.  A vision-based system would need to analyze the visual structure in sequences from hand-held or wearable cameras along path segments in order to answer two questions: which notional path or segment is being navigated, and where along a specific physical path, relative to start and end point, a person might be. We addressed the first of these questions in previous work \citep{Rivera-Rubio2013}.

\begin{figure}[t]
\begin{center}
\includegraphics[width=1\linewidth]{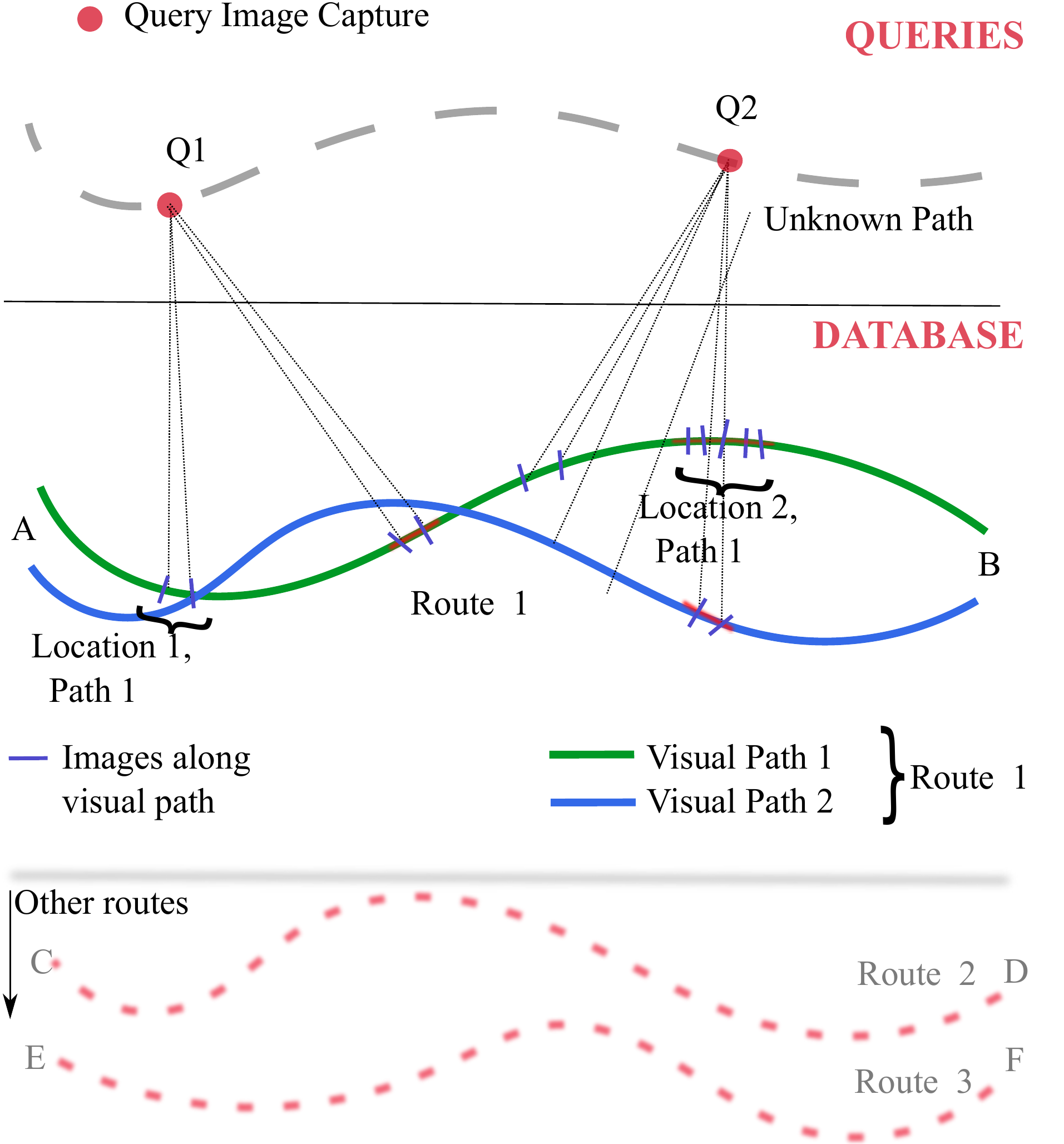}
\caption{The idea behind searching across data from navigators of the same physical path:  after navigating the space twice, Juan's visual path data (A, B) is indexed and stored in a database.  Mary enters the same space (\textit{unknown path}), and the images acquired as she moves are compared against the visual path of Juan, providing a journey-centric estimate of location. With many journeys collated, location can be inferred with respect to the pre-collected paths in the database.}
\label{fig:pathexample}
\end{center}
\end{figure}

Returning to the two-user scenario, let us assume that Juan has been the first to navigate along the path, and has collected a sequence of video frames during his successful navigation.  As Mary makes her way along the path, we wish to be able to associate the images taken by Mary with those taken by Juan (see Fig.~\ref{fig:pathexample}). The ability to do this allows us to locate Mary relative to the journey of Juan {\em from the visual data acquired by both}.  For only two users, this may seem an uninteresting thing to do.  However, imagine that this association is done between not two, but multiple users, and is applied to several physical paths that together form the navigable space of a building.  Achieving this association would enable some types of inference to be performed.  In particular:
\begin{itemize}
\item The visual path data would be a new source of data that could be used for location estimation;
\item The association of image locations would allow visual change detection to be performed over many journeys along the same route, made at different times;
\item Through non-simultaneous, many-camera acquisition, one could achieve more accurate mapping of a busy space, particularly where moving obstructions might be present;
\item  Visual object recognition techniques could be applied to recognize the nature of structures encountered along the route, such as exits, doorways and so on.
\end{itemize}

\begin{figure*}
\centering
\begin{subfigure}[b]{0.45\textwidth}
\centering
\includegraphics[width=0.7\linewidth]{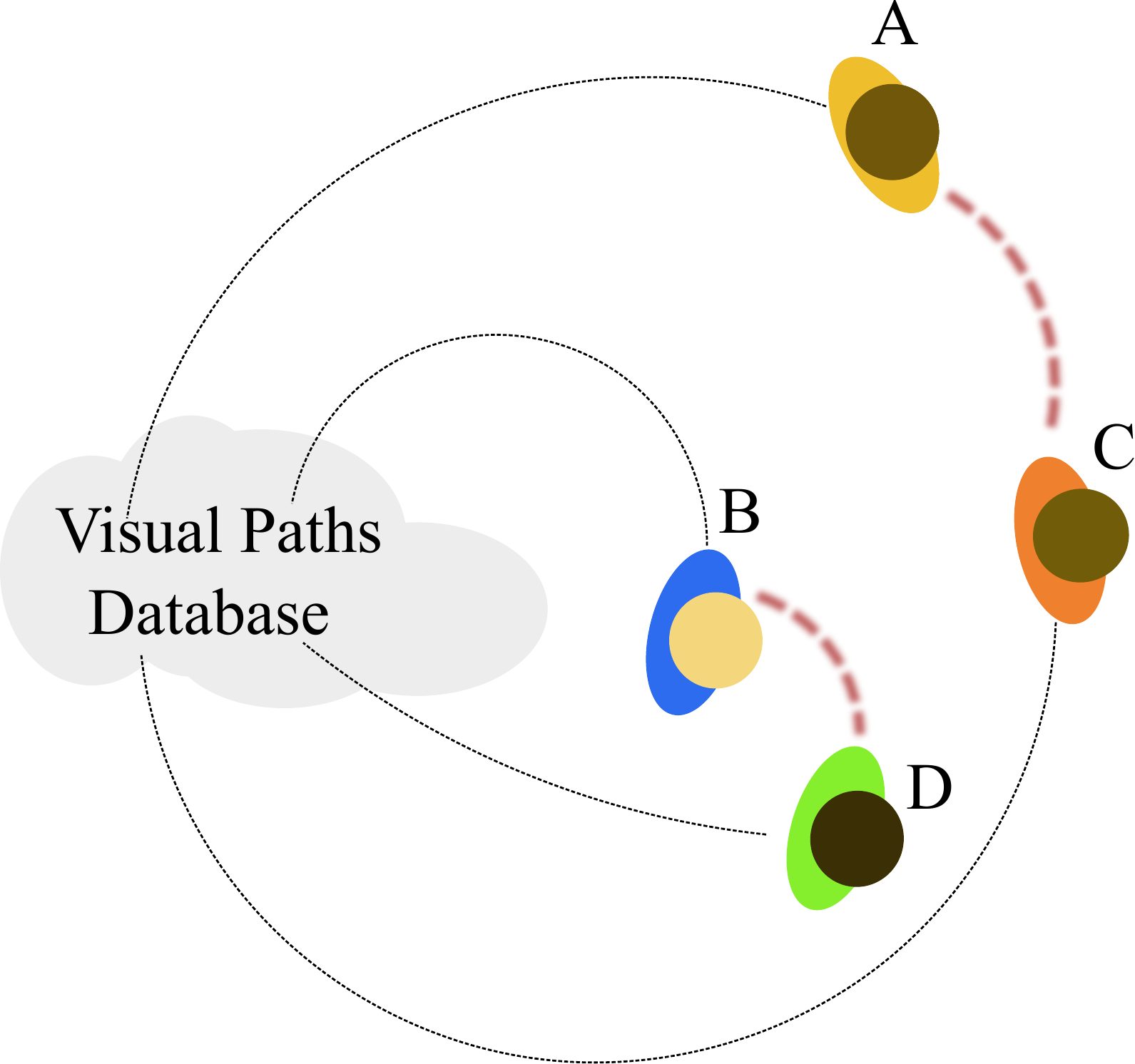}
\caption{}
\end{subfigure}
\hfill
\begin{subfigure}[b]{0.45\textwidth}
\centering
\includegraphics[width=\linewidth]{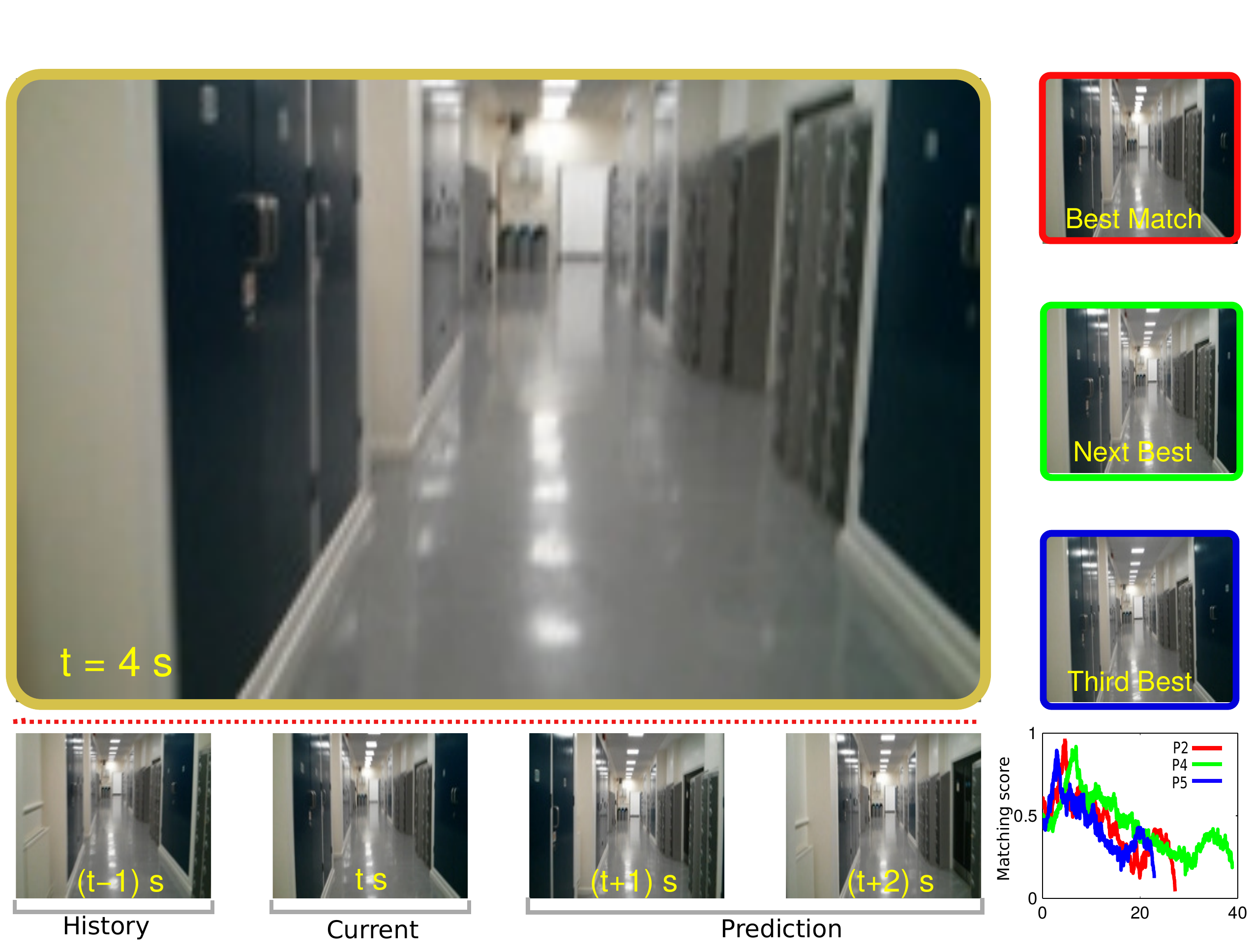}
\caption{}
\end{subfigure}
\caption{In (a), we illustrate the concept of using the visual path database to establish rough correspondence in users' locations through their visual experiences. Users (A,C) and (B,D) experience similar visual paths (see text for details).  In (b), the current view captured by a camera and views from the best match {\em paths} that have been captured through that space, to the immediate right.  The first four bottom panels show current, historical, and predicted images, based on the query from the best matching visual path.  The right, bottom image shows the similarity scores from other journeys taken along the same notional path.  The techniques that enable this type of match to be done are discussed in Section \ref{sec:Index}. }
\label{fig:ConceptAndHeadsUp}
\end{figure*}

Using continuously acquired images provides a new way for humans to interact with each other through establishing associations between the visual experiences that they have shared, independent of any tags that have been applied.  The concept is illustrated in Fig.~\ref{fig:ConceptAndHeadsUp}(a). In this diagram, four users are within the same region of a building; however, two pairs of users (A,C) and (B,D) are associated with having taken similar trajectories to each other. With a sufficient number of users, one could achieve a crowdsourcing of visual navigation information from the collection of users, notional paths and trajectories.

One intriguing possibility would be to provide information to visually-impaired users. For example, in an assistive system, the visual cues that sighted individuals experience along an indoor journey could be mined, extracting reliable information about position and objects (e.g. exit signs) that are of interest.  Whilst other sources of indoor positioning information, such as the locations of radio beacons, can aid indoor navigation, some visual cues are likely to be stable over long periods of time, and do not require extra infrastructure beyond that already commonly installed. Collecting distinctive visual cues over many journeys allows stable cues to be learned. Finally, in contrast to signal-based methods of location landmarks \citep{Wang2012}, the ``debugging'' of this type of navigation data -- i.e. images, or patches within images -- is uniquely human-readable: it can be done simply through human observation of what might have visibly changed along the path. Perhaps most compelling of all, visual path data can be acquired merely by a sighted user sweeping the route with a hand-held or wearable camera.

\section{Vision-Based Approaches to Navigation}
The current state-of-the-art methods for robot navigation make use of simple visual features and realistic robot motion models in order to map, then to navigate.  For human navigation, the challenge is slightly greater, due partly to the variability of human motion. Nevertheless, recent progress in simultaneous localization and mapping (SLAM) \citep{Newcombe2011} and parallel tracking and mapping (PTAM) \citep{Klein2009}  have yielded stunning results in producing geometric models of a physical, recovering geometry and camera pose simultaneously from hand-held devices. 

At the same time, being able to recognize certain objects whilst performing SLAM could improve accuracy, reducing the need for loop closure and allowing better -- more reliable -- self-calibration \citep{Salas-Moreno2013}.  Recognition pipelines in computer vision have recently taken great strides, both in terms of scalability and accuracy.  Thus, the idea of collaboratively mapping out a space through wearable or hand-held cameras is very attractive.

Appearance-based navigation, closely related to navigation, has been reported as one of many mechanisms used in biology, and has been explored by various groups in different animals (see, for example,
 \citep{collett2010desert,dombeck2010functional,fry2005look}). Appearance-based approaches can add to the information gained using SLAM-type algorithms. Indeed, in a robust system, we might expect several sources of localization information to be employed.  Consider, for example, outdoor navigation in cities: GPS can be combined with WiFi RSSI.  Doing so improves overall accuracy, because the errors in these two localization systems are unlikely to be highly correlated over relatively short length scales ($\approx 100$ m), and would only be trivially correlated (but highly) over longer distances. Localization systems often rely on motion models embedded into tracking algorithms, such as Kalman, extended Kalman \citep{Davison2003} filtering, or particle-filtering \citep{Pupilli2005}, to infer position. More recently, general purpose graphics processing units (GP-GPUs) have enabled camera position to be quickly and accurately inferred relative to a point cloud by registering whole images with dense textured models \citep{Newcombe2011}.

Anecdotal evidence and conversations with UK groups supporting visually-impaired people suggests that no {\em single} source of data or single type of algorithm will be sufficient to meet the needs of users who are in an unfamiliar space, or who might suffer from visual impairment. It is likely that a combination of sensors and algorithms is called for. 

\subsection{A Biological Perspective}
Research into the mechanisms employed by humans during pedestrian navigation suggests that multisensory integration plays a key role \citep{Panagiotaki2006}.  Indeed, studies into human spatial memory using virtual reality and functional neuroimaging \citep{burgess2002human,burgess2006spatial} suggest that the human brain uses a combination of representations to self-localize that might be termed as \textit{allocentric} and \textit{egocentric}. The egocentric representation supports identifying a location based on sensory patterns recognized from previous experiences in a given location. Allocentric representations use a reference frame that is independent of one's location.  The respective coordinate systems can, of course, be interchanged via simple transformations, but the sensory and cognitive processes underlying navigation in both cases are thought to be different.

The two forms of representation are typified by different types of cells, and, in some cases,  different neuronal signal pathways.  Within some mammals, such as  mice, it appears that a multitude of further sub-divisions of computational mechanisms lie behind location and direction encoding.  For example, in the hippocampus, there are at least four classes \citep{hartley2014space} of encoding associated with position and heading.  Hippocampal {\em place cells} display elevated firing when the animal is in a particular location \citep{dragoi2014selection}. The environmental cues that affect hippocampal place cells include vision and odour, so the inputs to these cells are not necessarily limited to any one type of sensory input.

Grid cells, on the other hand, show increased firing rates when the animal is present at a number of locations on a spatial grid; this suggests that some form of joint neuronal encoding is at work, and, indeed, there is some evidence that place cell responses arise through a combination of grid cells of different spacing \citep{moser2008place}. Boundary cells in the hippocampus appear to encode just that: the distance to the boundaries of a spatial region. This encoding seems to be relative to the direction the animal is facing but independent of the relation between the animal's head and body; they are therefore, examples of an allocentric scheme.  

In conclusion, biology seems to employ not only several sensory inputs to enable an organism to locate itself relative to the environment, but also different computational mechanisms.  The evidence of these multiple strategies for localization and navigation \citep{hartley2014space,poucet2014independence} motivates the idea for an appearance-based localization algorithm.

\section{The Dataset}
\label{sec:Dataset}
A total of 60 videos were acquired from 6 corridors of the RSM building at Imperial College London.  Two different devices were used. One was a LG Google Nexus 4 mobile phone running Android 4.4.2 ``KitKat''.  The video data was acquired at approximately 24-30 fps at two different acquisition resolutions, corresponding to $1280 \times 720$ and $1920\times 1080$ pixels.  The other device was a wearable Google Glass (2013 Explorer edition) acquiring data at a resolution of $1280 \times 720$, and a frame rate of around 30 fps.  A surveyor's wheel (Silverline) with a precision of 10 cm and error of $\pm 5\%$ was used to record distance, but was modified by connecting the encoder to the general purpose input/output (GPIO)  pins of a Raspberry Pi running a number of measurement processes.  The Pi was synchronized to network time using network time protocol (NTP), enabling synchronization with timestamps in the video sequence.  Because of the variable frame rate of acquisition, timestamp data from the video was used to align ground-truth measurements with frames.  This data is used to assess the accuracy of estimating positions, and not for any form of training.

In total, 3.05 km of data is contained in this dataset, at a natural indoor walking speed.  For each corridor, ten passes (i.e.\ 10 separate visual paths) are obtained; five of these are acquired with the hand-held Nexus, and the other five with Glass.  Table~\ref{tbl:Datasets} summarizes the acquisition.  As can be seen, the length of the sequences varies within some corridors; this is due to a combination of different walking speeds and/or different frame rates. Lighting also varied, due to a combination of daylight/nighttime acquisitions, and occasional windows acting as strong lighting sources in certain sections of the building.  Changes were also observable in some videos from one pass to another due to the presence of changes (path obstructions being introduced during a cleaning activity) and occasional appearances of people.

\begin{table}[h]
\caption{Detailed summary of the RSM dataset: $L$ length of the corridors and $Fr$ number of video frames.  The single frames are representative images from hand-held videos of selected corridors of the RSM building at Imperial College London.  The dataset includes both hand-held and wearable camera examples, all containing ground-truth location relative to distance traversed along labelled paths. The grand totals are: $L$ = 3,042 km and \#Fr = 90,302 frames.}
\label{tbl:Datasets}
    \begin{tabular}{L{0.78cm} C{0.9cm} C{0.9cm} C{0.9cm} C{0.9cm} C{0.9cm} C{0.9cm}  }
    &
	 \begin{minipage}{0.083\textwidth}
      			\includegraphics[width=0.8\linewidth]{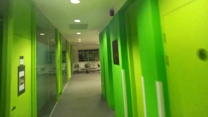}
			   \end{minipage} 
			   & 
			   \begin{minipage}{0.083\textwidth}
      			\includegraphics[width=0.8\linewidth]{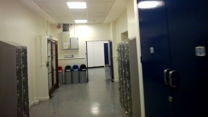}
			   \end{minipage} 
			   & 
			   \begin{minipage}{0.083\textwidth}
      			\includegraphics[width=0.8\linewidth]{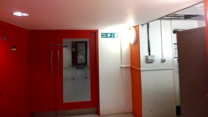}
			   \end{minipage} 
			   & 
			   \begin{minipage}{0.083\textwidth}
      			\includegraphics[width=0.8\linewidth]{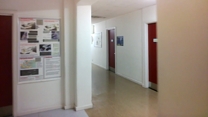}
			   \end{minipage} 
			   & 
			   \begin{minipage}{0.083\textwidth}
      			\includegraphics[width=0.8\linewidth]{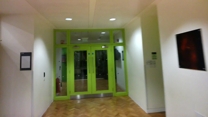}
			   \end{minipage} 
			   & 
			   \begin{minipage}{0.083\textwidth}
      			\includegraphics[width=0.8\linewidth]{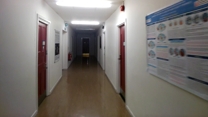} 
			   \end{minipage} 
			   \\
			    \small{$\bar{L}$(m)} & 57.9 & 31.0 & 52.7 & 49.3 & 54.3 & 55.9 \\
			    \small{$\sum{L}$} & 585.6 & 312.4 & 524.2 & 497.7 & 562.0 & 560.4 \\
			    \hline \hline
			    $\bar{Fr}$ & 2157 & 909 & 1427 & 1583 & 1782 & 1471 \\
			      {$\sum{Fr}$} & 19379 & 9309 &14638 & 15189 & 16823 & 14964 \\

    \end{tabular}
\end{table}

In total, more than 90,000 frames of video were labelled with positional ground-truth. The dataset is publicly available for download at \url{http://rsm.bicv.org} \citep{Rivera-Rubio2014}.

\section{Methods: Indexing}
\label{sec:Index}
\begin{figure}
\begin{center}
\includegraphics[width=\linewidth]{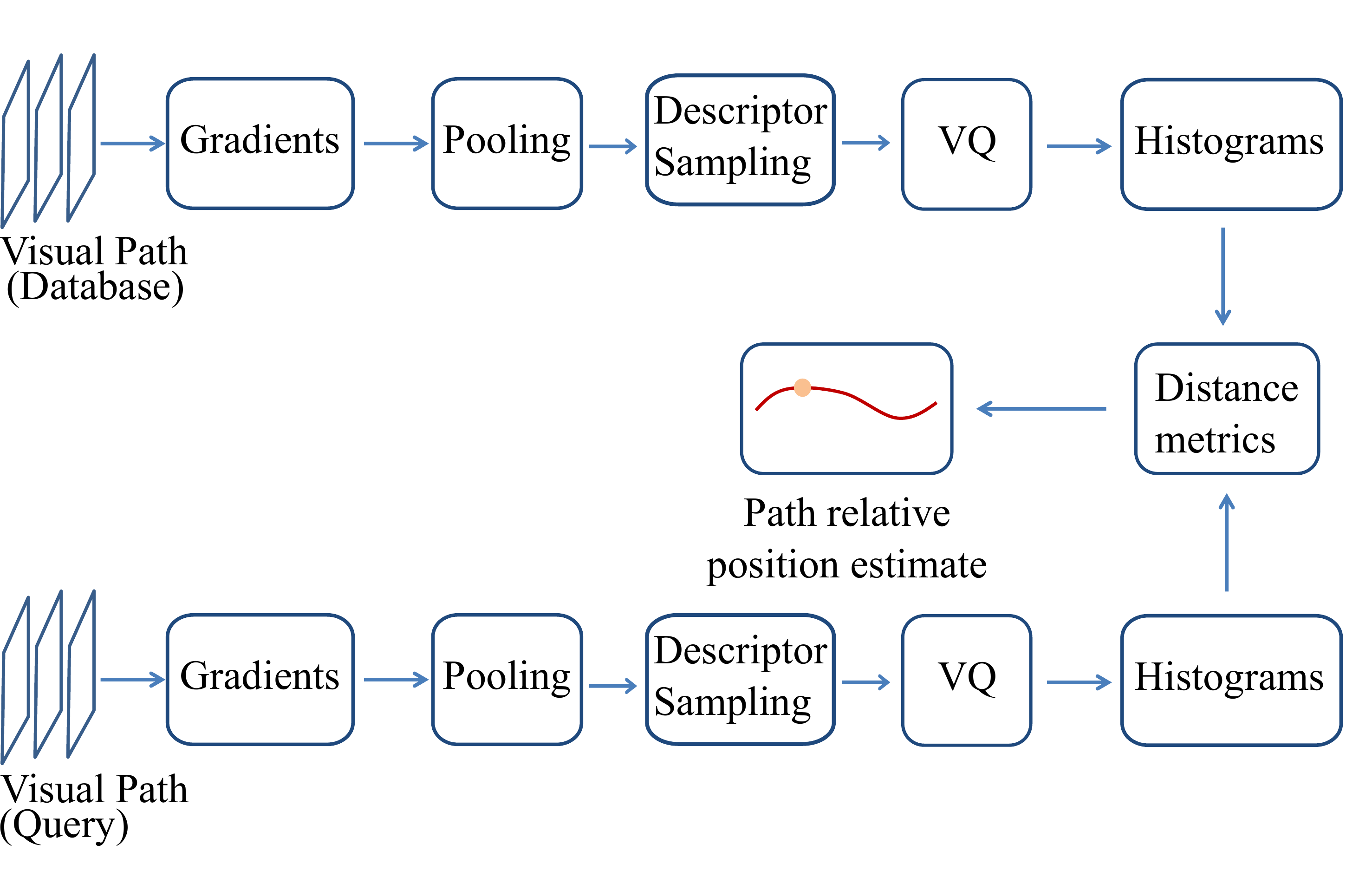}
\caption{This diagram illustrates the stages in processing the  sequences in the database and for the queries.  This diagram does not show the process behind the estimation of ground-truth for the experiments, which is described separately in Section \ref{sec:Dataset}.  Variants of the gradient operators, pooling operators, quantization and distance metrics are described in Section \ref{sec:Index}.} 
\label{fig:FigPipeline}
\end{center}
\end{figure}

We evaluated the performance of different approaches to query images taken from one visual path against others stored in the database.  In order to index and query visual path datasets, we used the steps illustrated in Fig. \ref{fig:FigPipeline}.  The details behind each of the steps (e.g. gradient estimation, spatial pooling) are described in the remainder of this section.  They include techniques that operate on {\em single} frames as well as descriptors that operate on {\em multiple} frames, at the frame level and at the patch level. All the performance evaluation experiments were carried out at low-resolution (208$\times $117 pixels) versions of the sequences, keeping bandwidth and processing requirements small.

\subsection{Frame-Level Descriptor}
\begin{figure}
\begin{center}
\includegraphics[width=\linewidth]{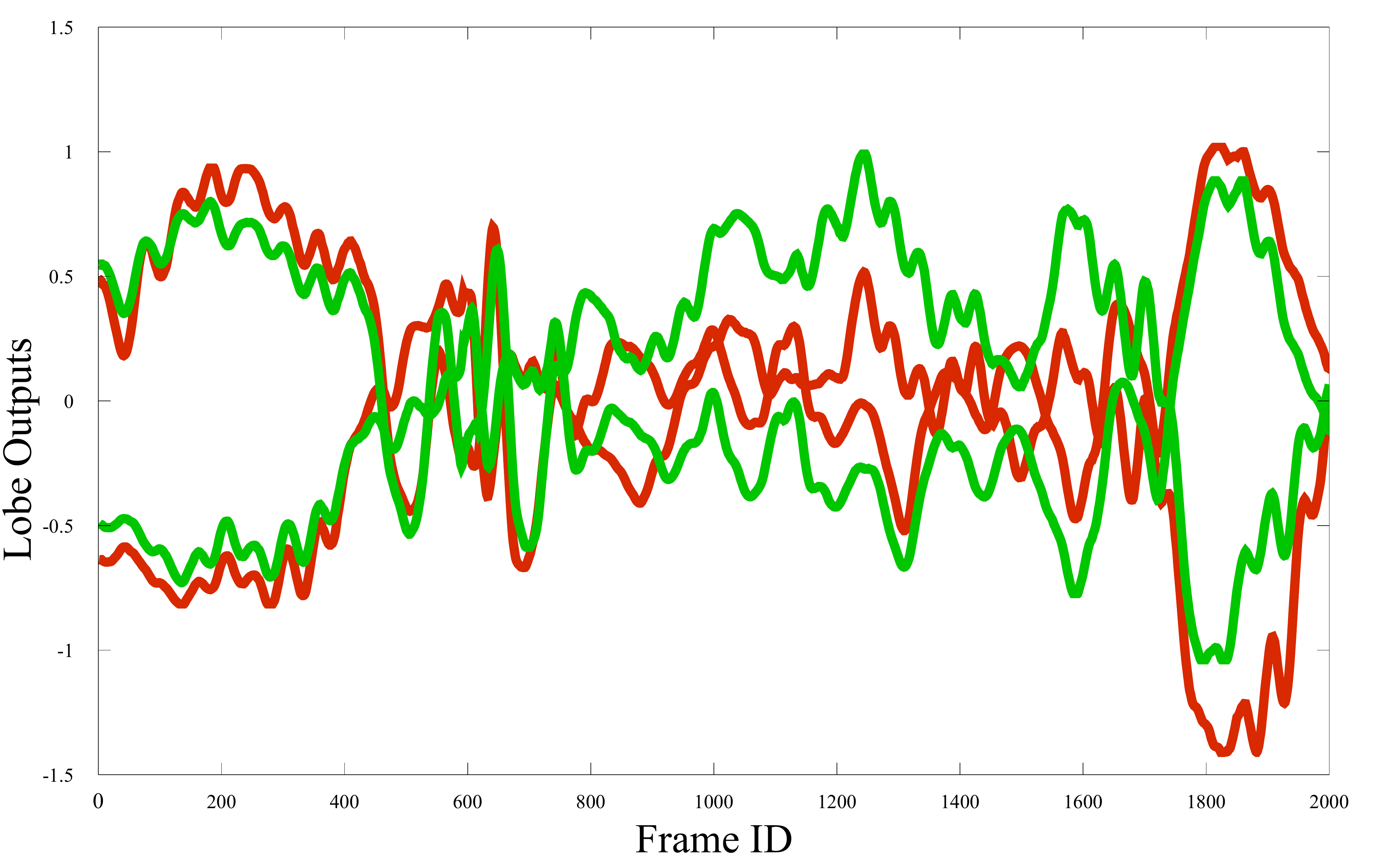}
\caption{Four (of 144) representative signals acquired from a visual path; these signals encode changes in red and green channels as a user moves through space.  The collection of signal traces at one point in time can be used to build a simple frame-level space-time descriptor: LW\_COLOR. The signal amplitudes are spatially pooled temporal and spatial gradient intensities.}
\label{fig:Traces}
\end{center}
\end{figure}

Based on the use of optical flow in motion estimation~\citep{Weickert2006} and space-time descriptors in action recognition \citep{Wang2009} we estimated in-plane motion vectors using a simple approach.  We first applied derivative filters along $(x,y,t)$ dimensions, yielding a 2D$+t$, i.e. spatio-temporal, gradient field.  To capture variations in chromatic content from the visual sequence, we computed these spatio-temporal gradients separately for each of the three RGB channels of the pre-processed video sequences.  This yielded a $3\times 3$ matrix at each point in space.  Temporal smoothing was applied along the time dimension, with a support of 11 neighbouring frames. Finally, the components of the matrix were each averaged (pooled) over 16 distinct spatial regions, not very dissimilar to those to be described later in this paper.   For each visual path, this yielded 144 signals, of length approximately equal to the video sequences.  An illustration of the time series for one visual path is shown in Fig.~\ref{fig:Traces}.

At each point in time, the values over the 144 signal channels are also captured into a {\it single} space-time descriptor per frame: LW\_COLOR.  Our observations from the components of this descriptor are that a) relative ego-motion is clearly identifiable in the signals; b) stable patterns of motion may also be identified, though changes in the precise trajectory of a user could also lead to perturbations in these signals, and hence to changes in the descriptor vectors. Minor changes in trajectory might, therefore, reduce one's ability to match descriptors between users.  These observations, together with the possibility of partial occlusion, led us to the use of {\em patch} based descriptors, so that multiple descriptors would be produced for each frame. These are introduced next.

\subsection{Patch-Level Descriptors}
\label{subsec:Patch}
The patch descriptors can be further divided into two categories: those produced from patches of single frames, and those that are based on patches acquired over multiple frames; the latter are {\em space-time} patch descriptors.  We explored two distinct single-frame descriptors, and three distinct space-time descriptors.  We first describe the single-frame descriptors.

\subsubsection{Spatial Patch Descriptors (single-frame)} 
The spatial patch descriptors consist of the Dense-SIFT descriptor  \citep{Lowe1999,Lazebnik2006,Vedaldi2008} and a tuned, odd-symmetric Gabor-based descriptor.  The SIFT descriptor was calculated by dense sampling of the smoothed estimate of $\vec{\nabla}f(x,y;\sigma)$ where $f(x,y;\sigma)$ represents the scale-space embedding of image $f(x,y)$ within a Gaussian scale-space at scale $\sigma$.  We used a standard implementation of dense SIFT from VLFEAT~\citep{Vedaldi2008} with scale parameter, $\sigma \approx 1$, and with a stride length of 3 pixels. This yielded around $2,000$ descriptors per frame, each describing a patch of roughly $10 \times 10$ pixels in the frame.

\begin{figure}[h]
\centering
\includegraphics[width=\linewidth]{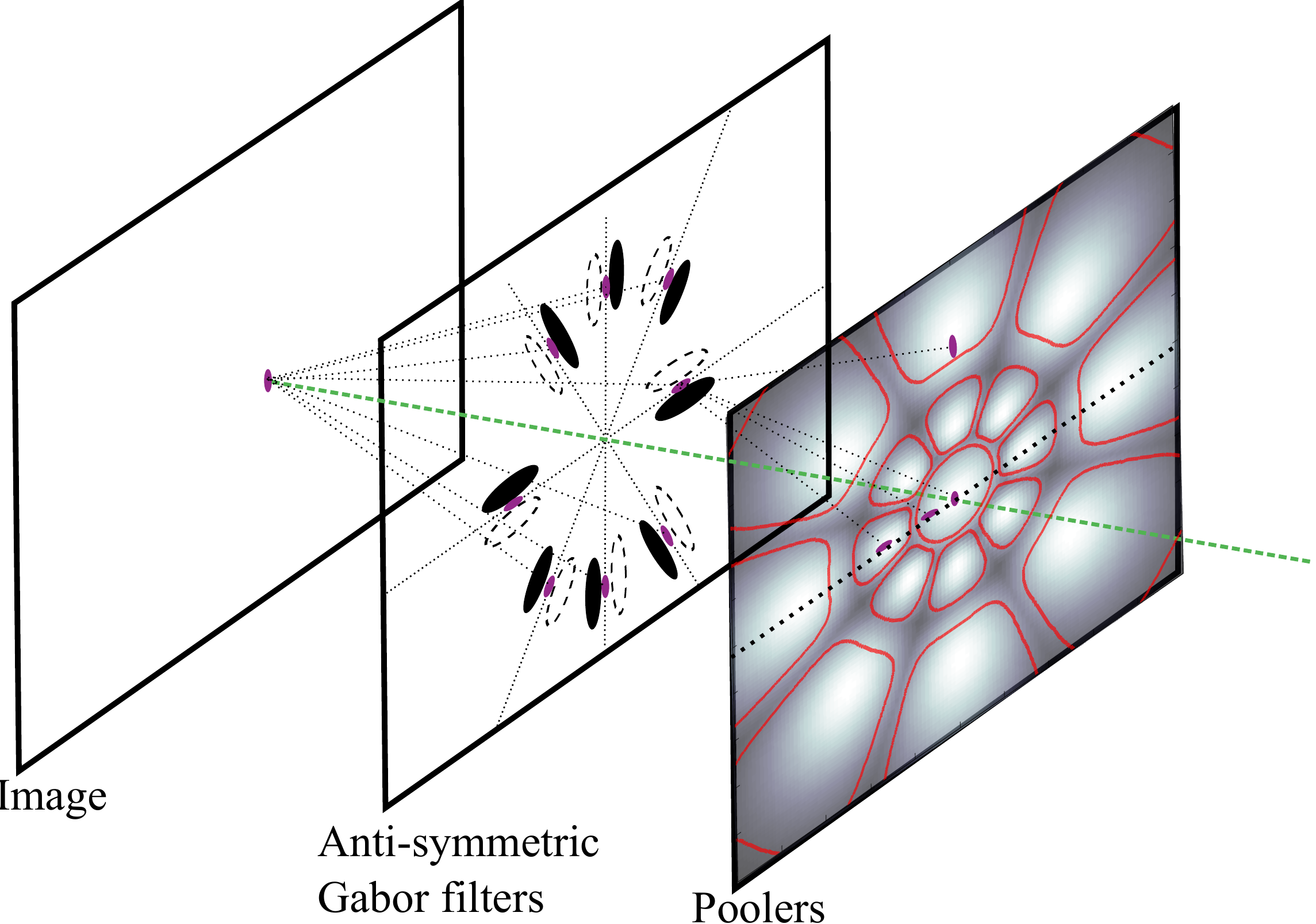}
\caption{Orientation selective masks are applied to the video frames by spatial convolution.  The outputs of the masks are collected over space using the pooling functions. The outputs of the poolers are sub-sampled over the image plane to produce descriptors for indexing.  See text for further details.}
\label{fig:IsoPool}
\end{figure}

We compared these with another single-frame technique devised in our lab: we used filters that we previously tuned on PASCAL VOC data \citep{Everingham2009} for image categorization.  These consisted of $8$-directional, 9 $\times$ 9 pixels spatial Gabor filters ($k = 1,...,8$; $\sigma = 2$). Each filter gives rise to a filtered image plane, denoted $\mathbf{G}_{k,\sigma}$.  For each plane, we applied spatial convolution ($\ast$) with a series of pooling functions:

\begin{equation}
\mathbf{G}_{k,\sigma} \ast {\Phi}_{m,n}
\label{eq:conv}
\end{equation}

where ${\Phi}_{m,n}$ is computed by spatial sampling of the pooling function:

\begin{equation}
\Phi(x,y;m,n) = \exp\left \lbrace -\alpha \left [\log_e \left ( \frac{x^2+y^2}{d_n^2}\right ) \right ]^2 - \beta | \theta-\theta_m | \right \rbrace
\label{eq:pool1}
\end{equation}

\noindent with $\alpha = 4$ and $\beta = 0.4$. The values of $m = 0,...,7$ and $n = 0,1,2$ were taken to construct 8 regions at angles $\theta_m = m\frac{\pi}{4}$ for each of two distances $d_1 = 0.45$, $d_2 = 0.6$ away from the centre of a spatial pooling region in the image plane.  For the central region, corresponding to $m=0$, there was no angular variation, but a log-radial exponential decay. This yielded a total of  17 spatial pooling regions. The resulting $17 \times 8$ fields were sub-sampled to produce a dense 136-dimensional descriptors, each representing an approximately $10 \times 10$ pixels image region. This resulted in, again, approximately 2,000 descriptors per image frame after the result of Eq.~(\ref{eq:conv}) is sub-sampled. This is illustrated in Fig.~\ref{fig:IsoPool}.

\subsubsection{Space-Time Patch Descriptors}
Given the potential richness available in the capture of space-time information, we explored three distinct approaches to generate space-time patch descriptors.  These approaches all lead to multiple descriptors per frame, and all take into account neighbouring frames in time when generating the descriptor associated with each patch.  Additionally, all three densely sample the video sequence.  The three methods are i) HOG 3D, introduced by \citet{Klaser2008}; our space-time, antisymmetric Gabor filtering process (ST\_GABOR); and iii) our spatial derivative, temporal Gaussian (ST\_GAUSS) filter.

\begin{enumerate}
\item The HOG 3D descriptor (HOG3D) \citep{Klaser2008} was introduced to extend the very successful two-dimensional histogram of oriented gradients technique \citep{Dalal} to space-time fields, in the form of video sequences.  HOG 3D seeks computational efficiencies by smoothing using box filters, rather than Gaussian spatial or space-time kernels.  This allows three-dimensional gradient estimation across multiple scales using the {\em integral video} representation, a direct extension of the integral image idea \citep{Viola2001}.  The gradients from this operation are usually performed across multiple scales.  We used the dense HOG 3D option from the implementation of the authors \citep{Klaser2008}, and the settings yielded approximately 2,000 descriptors per frame of video.

\item Space-time Gabor (ST\_GABOR) functions have been used in activity recognition, structure from motion and other applications \citep{Bregonzio2009}.  We performed one dimensional convolution between the video sequence and three one-dimensional Gabor functions along either one spatial dimension i.e.\ $x$ or $y$, or along $t$.  The one-dimensional convolution is crude, but appropriate if the videos have been downsampled. The spatial extent of the Gabor was set to provide one complete cycle of oscillation over approximately 5 pixels of spatial span, both for the $x$ and $y$ spatial dimensions. The filter for the temporal dimension was set to provide temporal support and one oscillation over approximately 9 frames.  We also explored symmetric Gabor functions, but found them less favourable in early performance tests.

After performing three separate filtering operations, each pixel of each frame was assigned a triplet of values corresponding to the result of the each filtering operation.  The three values were treated as being components of a 3D vector.  Over a spatial extent of around $16 \times 16$ pixels taken at the central frame of the 9-frame support region, these vectors contribute weighted votes into 13 histogram bins according to their azimuth and elevations, with the weighting being given by the length of the vector.  The votes were also partitioned into 17 regions according to the approximate spatial lobe pattern illustrated in Fig.~\ref{fig:IsoPool}, yielding a 221-dimension descriptor.

\item A final variant of space-time patch descriptor was designed.  This consisted of spatial derivatives in space, combined with smoothing over time (ST\_GAUSS).  In contrast to the strictly one-dimensional filtering operation used for the space-time Gabor descriptor, we used two $5\times 5$ gradient masks for the $x$ and $y$ directions based on derivatives of Gaussian functions, and an 11-point Gaussian smoothing filter in the temporal direction with a standard deviation of 2.  8-directional quantization was applied to the angles of the gradient field, and a weighted gradient magnitude voting process was used to distribute votes across the 8 bins of a 136-dimensional descriptor.  Like the ST\_GABOR descriptor, pooling regions were created, similar to those shown in Fig.~\ref{fig:IsoPool}.
\end{enumerate}

\subsection{Frame-Level Encoding} \label{sec:encodings} 
Our initial conjecture was that whole frames from a sequence could be indexed compactly, using the single-frame descriptor (LW\_COLOR).  This was found to lead to disappointing performance (see Section~\ref{sec:Performance}). For the case of many descriptors-per-frame i.e.\ descriptors that are patch-based, we have the problem of generating around 2,000 descriptors per frame, if dense sampling is used.  Thus, we applied vector quantization (VQ) to the descriptors, then used histograms of VQ descriptors, effectively representing each frame as a histogram of words \citep{Csurka2004}. The dictionary was always built by excluding the entire journey from which queries are to be taken.  

Two different approaches to the VQ of descriptors were taken, one based on standard $k$-means, using a Euclidean distance measure (hard assignment, ``HA''), and one corresponding to the Vector of Locally Aggregated Descriptors (VLAD) \citep{jegou2010aggregating}. For VLAD, a $k$-means clustering was first performed. For each descriptor, sums of residual vectors were used to improve the encoding.  Further advances to the basic VLAD, which include different normalizations and multiscale approaches, are given by \cite{Arandjelovic}. To compare encodings, either $\chi^2$ or Hellinger distance metrics \citep{Vedaldi2012} were used to retrieve results for HA and VLAD encoding approaches respectively. Distance comparisons were performed directly between either hard assigned Bag-of-Words (BoW) or VLAD image encodings arising from collections of descriptors for each frame.

\section{Experiments and Results: Performance Evaluation}
\label{sec:Performance}
The methods for a) describing spatial or space-time structure, b) indexing and comparing the data are summarized in Table~\ref{tbl:Methods}. The choice of parameters was selected to allow a) as consistent a combination of methods as possible, allowing fair comparisons of the effect of one type of encoding or spatio-temporal operator to be isolated from others b) to select parameter choices close to other research in the area, e.g.\ for image categorization, dictionary sizes of $\approx 256$ and $\approx 4000$ words are common. 

\begin{table}
\caption{A summary of the different encoding methods and their relationships to different descriptors. The number of elements of each descriptor is also reported (\textbf{Dim}).}
\label{tbl:Methods}
\small
\centering
    \begin{tabular}{l p{0.5cm} p{0.9cm} p{0.5cm} p{1.5cm} c }
    \hline
    \textbf{Method}              & \textbf{ST} & \textbf{Dense} & \textbf{Dim}  & \textbf{Encoding}      & \textbf{Metric}    \\ \hline
    \multirow{2}{*}{SIFT} & \multirow{2}{*}{No} & \multirow{2}{*}{Yes} & \multirow{2}{*}{128}     & HA-4000  &   $\chi^2$    \\ \cline{5-6}
    ~                   & ~      & ~           & ~      & VLAD-256 & Hellinger \\ \hline
    \multirow{2}{*}{SF-GABOR} & \multirow{2}{*}{No}               & \multirow{2}{*}{Yes}  & \multirow{2}{*}{136}  & HA-4000  & $\chi^2$      \\  \cline{5-6}
    ~                   & ~      & ~           & ~      & VLAD-256 & Hellinger \\ \hline
    LW-COLOR           & Yes & No & 144           & N/A       \\ \hline
    \multirow{2}{*}{ST\_GABOR}          & \multirow{2}{*}{Yes}              & \multirow{2}{*}{Yes}  & \multirow{2}{*}{221}  & HA-4000  & $\chi^2$      \\  \cline{5-6}
    ~                   & ~      & ~           & ~      & VLAD-256 & Hellinger \\ \hline
    \multirow{2}{*}{ST\_GAUSS}           & \multirow{2}{*}{Yes}              & \multirow{2}{*}{Yes}  & \multirow{2}{*}{136}  & HA-4000  & $\chi^2$      \\  \cline{5-6}
    ~                   & ~      & ~           & ~      & VLAD-256 & Hellinger \\ \hline
    \multirow{2}{*}{HOG3D}               & \multirow{2}{*}{Yes}              & \multirow{2}{*}{Yes} & \multirow{2}{*}{192}   &  HA-4000  & $\chi^2$      \\  \cline{5-6}
    ~                   & ~       & ~          & ~      & VLAD-256 & Hellinger \\ \hline
    \end{tabular}
    \normalsize
\end{table}

\subsection{Error Distributions} \label{subsec:errdistr}
\begin{figure}[h]
\centering
\includegraphics[width=\linewidth]{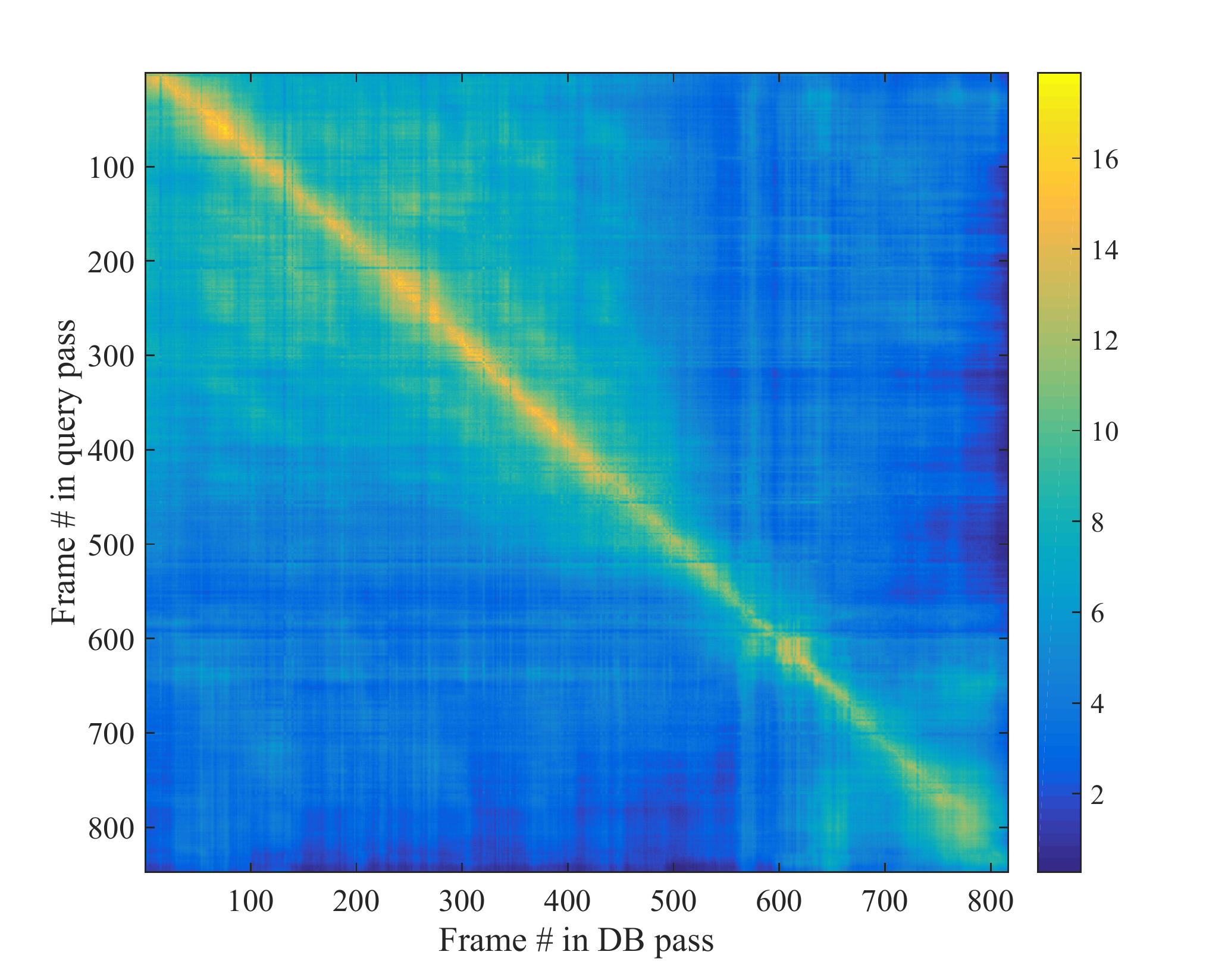}
\caption{Example of a $\chi^2$ kernel produced by hard assignment and using the SF\_GABOR descriptors when querying with pass 1 of corridor 2 against a database comprised of passes 2-10.}
\label{fig:kernel}
\end{figure}

Error distributions allow us to quantify the accuracy of being able to estimate {\em locations} along physical paths within the RSM dataset described in Section~\ref{sec:Dataset}. To generate the error distributions, we did the following: 

We started by using the kernels calculated in Section~\ref{sec:encodings}. One kernel is shown in Fig.~\ref{fig:kernel}, where the rows represent each frame from the query pass, and the columns represent each frame from one of the remaining database passes of that corridor. The values of the kernel along a row represent a ``score'' between a query and different database frames. In this experiment, we associated the position of the best match to the query frame, and calculated the error between this and the ground truth $\epsilon$, in cm. In order to characterise the reliability of such scores, we performed bootstrap estimates of error distributions using 1 million trials. The distribution of the errors gives us a probability density estimate, from which we can get the cumulative distribution function (CDF) $P(x \leq |\epsilon|)$. The outcome is shown in Fig.~\ref{fig:CDFAll}, where only the average across all the randomized samples is shown.

By permuting the paths that are held in the database and randomly selecting queries from the remaining path, we were able to assess the error distributions in localization.  Repeated runs with random selections of groups of frames allowed the variability in these estimates to be obtained, including that due to different numbers of paths and passes being within the database. 

If we consider the idea of crowdsourcing journey information from many pedestrian journeys through the same corridors, this approach to evaluating the error makes sense:  all previous journeys could be indexed and held in the database; new journey footage would be submitted as a series of query frames (see Fig.~\ref{fig:pathexample}).    

\subsection{Localization Error vs Ground-Truth Route Positions}
\begin{figure}[!ht]

\begin{subfigure}{\linewidth}
\centering
\includegraphics[width=0.95\textwidth]{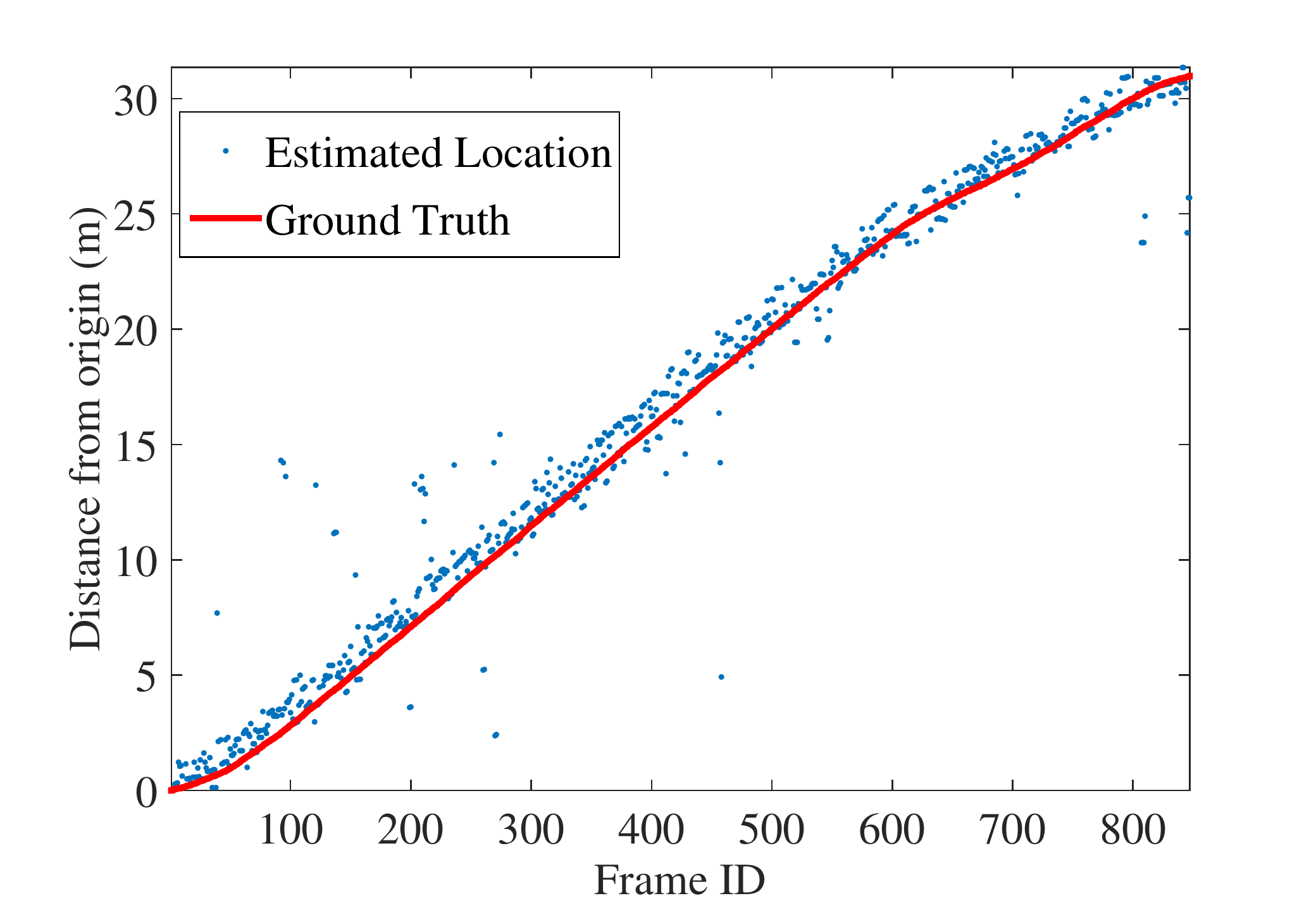}
\caption{Corridor 3 using Pass 2 acquired with LG Nexus 4. Results for the best spatio-temporal method, ST\_GABOR.}\label{fig:error_best}
\end{subfigure}%

\begin{subfigure}{\linewidth}
\centering
\includegraphics[width=0.95\textwidth]{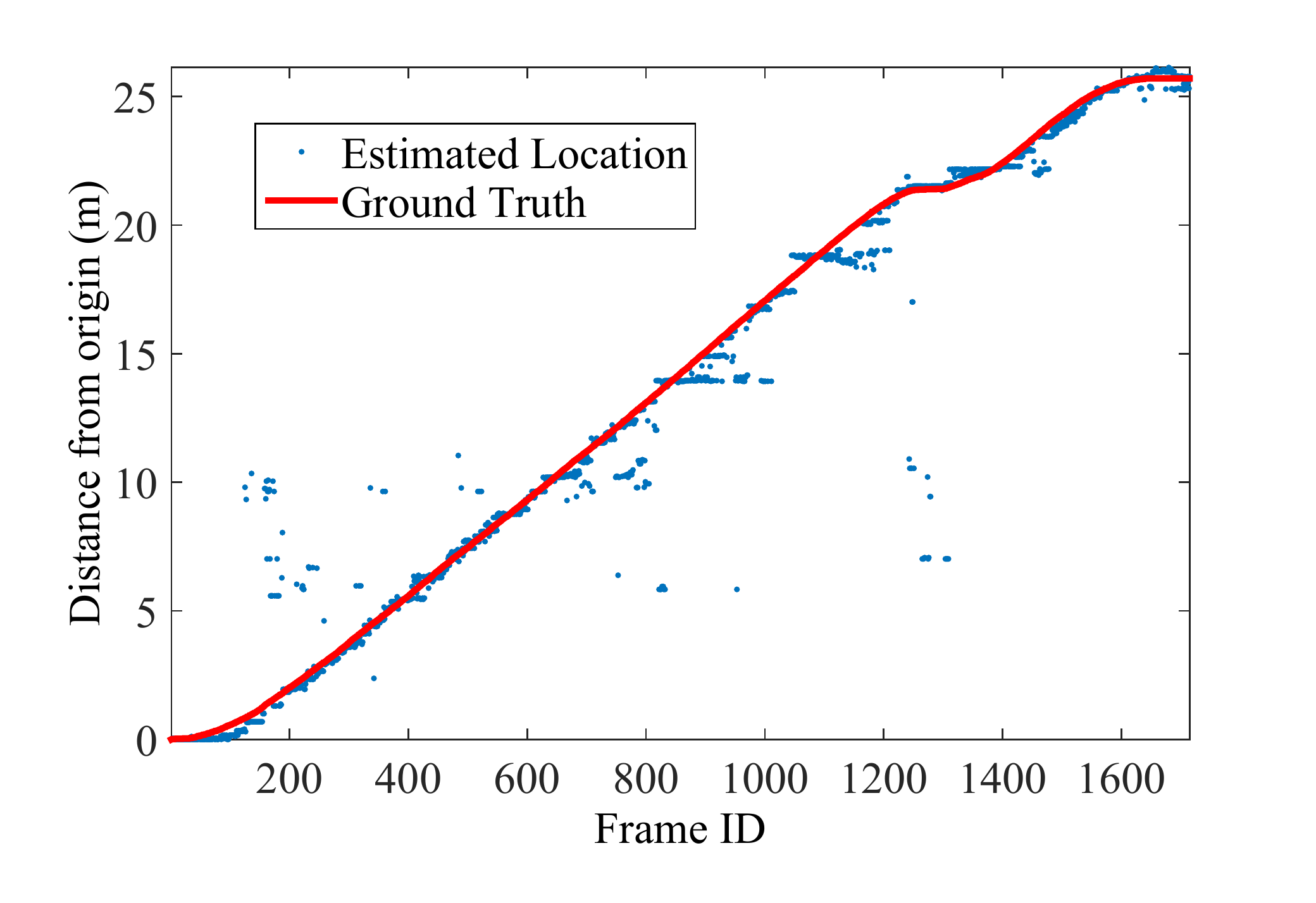}
\caption{Corridor 3 using Pass 6 acquired with Google Glass. Results for the best single-frame method, SF\_GABOR.}\label{fig:error_best}
\end{subfigure}%

\begin{subfigure}{\linewidth}
\centering
\includegraphics[width=0.95\textwidth]{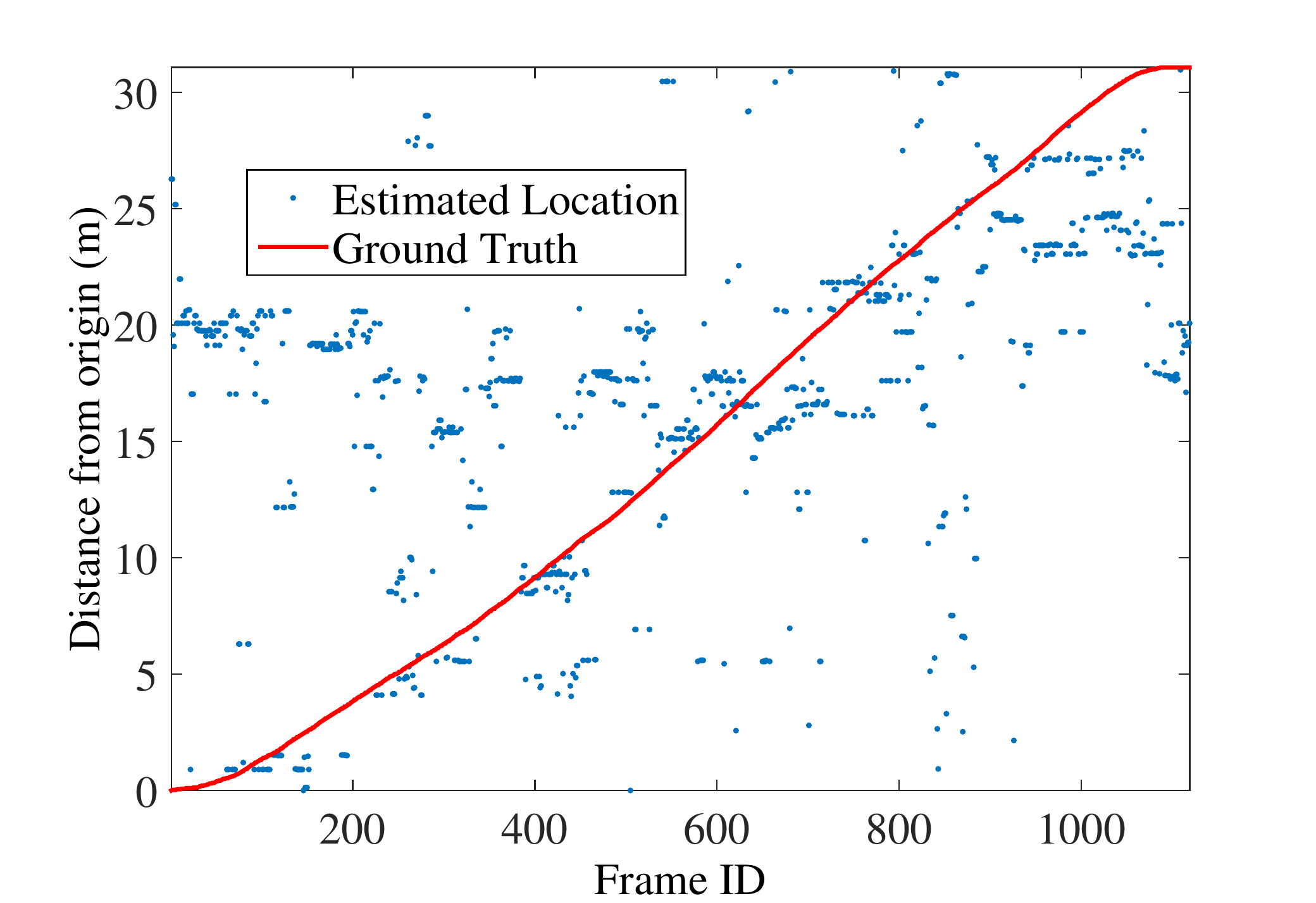}
\caption{Corridor 2 using Pass 6 acquired with Google Glass. Results for the worst overall method, LW\_COLOR.}\label{fig:error_worst}
\end{subfigure}%
\caption{Estimated location vs. ground truth. Illustrative examples of good/bad location estimation performance. a) Uses the best descriptor and a single-device dataset, b) uses the best descriptor and a cross-device dataset and c) uses the worst descriptor, and a multiple-device dataset.}
\label{fig:Glass}
\end{figure}

As described in the previous section, by permuting the database paths and selecting, randomly, queries from the remaining path that was left out in the dictionary creation, we can assess the errors  in localization along each corridor for each pass, and calculate, also, the average error in localization on a per-corridor basis, or per-path basis.  For these, we used the ground-truth information acquired as described in Section \ref{sec:Dataset}.  Fig.~\ref{fig:Glass} provides some examples of the nature of the errors, showing evidence of those locations that are often confused with each other.  As can be seen, for the better method (top trace of Fig.~\ref{fig:Glass}) whilst average errors might be small, there are, occasionally, large errors due to poor matching (middle trace). Errors are significantly worse for queries between different devices (see Fig.~\ref{fig:Glass}(c)). 

Note that we did not use any tracking algorithms, and so there is no motion model or estimate of current location given the previous one. Incorporating a particle filter or Kalman filter should reduce the errors, particularly where there are large jumps within small intervals of time. This deliberate choice allows us to evaluate the performance of different descriptor and metric choices independently.

\subsection{Performance Summaries}
\begin{figure}[h]
\centering
\includegraphics[width=\linewidth]{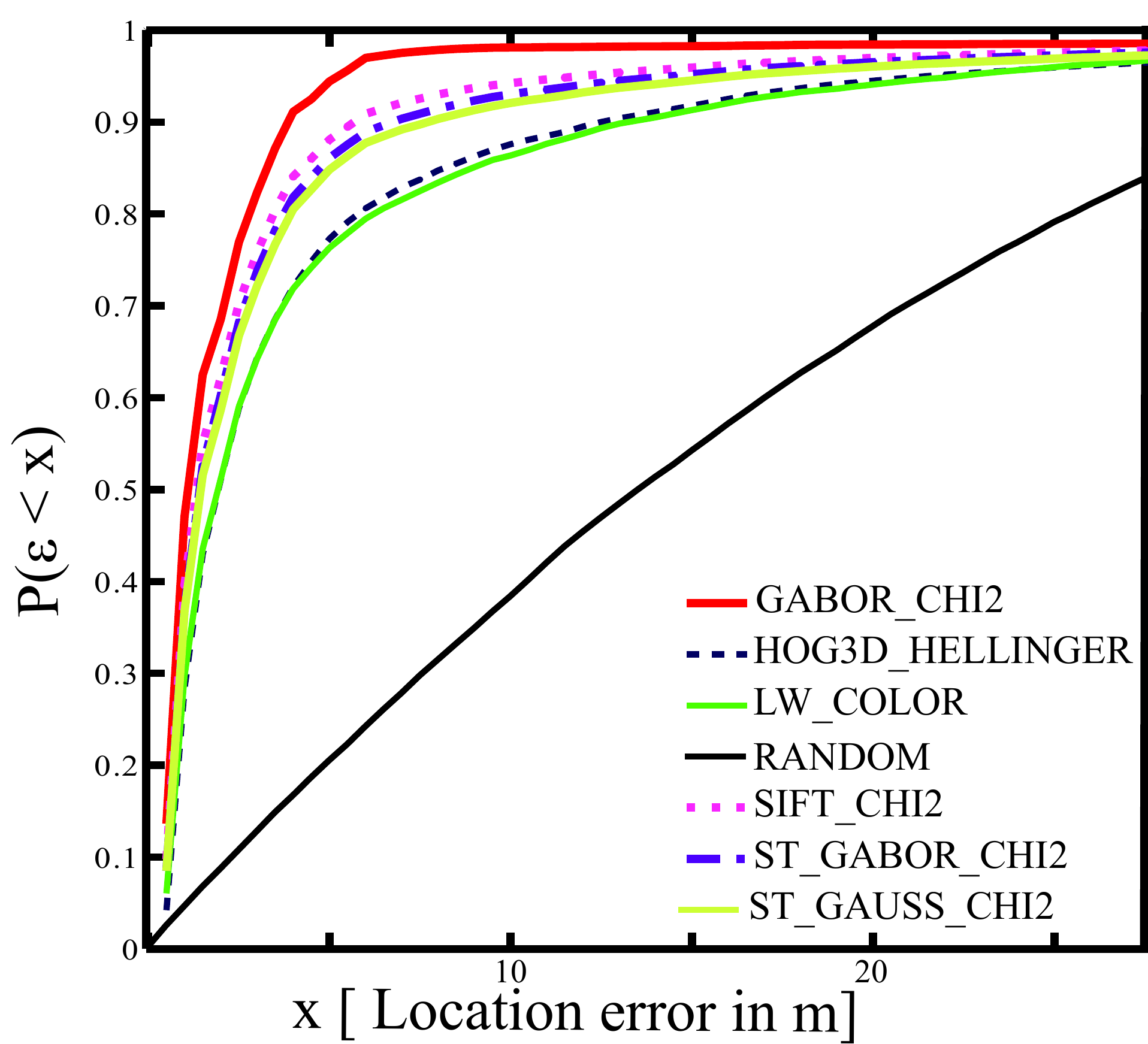}
\caption{Comparison of best performing version of all methods. The average CDFs were computed from the distributions generated after the random sampling described in Section \ref{subsec:errdistr}.  The results for a random test (RANDOM) were introduced as a ``sanity check''.}\label{fig:CDFAll}
\end{figure}

We calculated the average of the absolute positional error (in cm) and the standard deviation of the absolute positional error in a subset of the complete RSM dataset (Table~\ref{Table:summaries}). We  used a leave-one-journey-out approach (all the frames from an entire journey are excluded from the database).  Using bootstrap sampling, we also estimated the cumulative density functions of the error distributions in position, which are plotted in Fig.~\ref{fig:CDFAll}. The variability in these curves is not shown, but is summarized in the last two columns of Table~\ref{Table:summaries} through the area-under-curve (AUC) values. In the best case (SF\_GABOR), AUCs of the order of $~96\%$ would mean errors generally below 2 m; in the worst (HOG3D), AUCs $\approx$ 90\% would mean errors of around 5 m.  These  mean absolute error estimates are obtained as we permute the queries, the dictionary and the paths in the database. 

\begin{table}[h]
\caption{Summaries of average absolute positional errors and standard deviation of positional errors for different descriptor types and for different encoding methods (labelled by the corresponding metric used: $\chi^2$ for HA and Hellinger for VLAD). $\mu_{\epsilon}$ is the average absolute error, and $\sigma_{\epsilon}$ is the standard deviation of the error in cm. Single device case and in bold: best and worst AUC.}
\small
    \begin{tabular}{ c c c c c c }
    \hline
     \multirow{2}{*}{\bf Method} &  \multirow{2}{*}{\bf Metric} & \multicolumn{2}{c}{Error summary (cm)} & \multicolumn{2}{c}{AUC (\%)}\\ \cline{3-4}    
     \cline{5-6}
    & & $\mu_{\epsilon}$ & $\sigma_{\epsilon}$ & Min & Max \\ \hline

    SF\_GABOR & $\chi^2$       & 130.6              & 38.8             & 96.40   & \textbf{96.75}   \\ \hline
    SF\_GABOR & Hellinger      & 135.1              & 46.5              & 96.29   & 96.71   \\ \hline
    ST\_GAUSS & $\chi^2$       & 135.4              & 44.1              & 93.61   & 94.30   \\ \hline
    ST\_GAUSS & Hellinger      & 144.1              & 52.4              & 92.69   & 93.47   \\ \hline
    ST\_GABOR & $\chi^2$       & 235.9              & 86.3              & 93.97   & 94.66   \\ \hline
    ST\_GABOR & Hellinger      & 179.5              & 62.3              & 93.98   & 94.60   \\ \hline

SIFT     & $\chi^2$       & 137.5              & 46.3              & 94.57   & 95.14   \\ \hline
    SIFT     & Hellinger      & 132.7              & 41.4              & 94.34   & 94.95   \\ \hline

    HOG3D    & $\chi^2$       & 419.6              & 133.3              & \textbf{90.89}   & 91.83   \\ \hline
    HOG3D    & Hellinger      & 366.5              & 120.3              & 91.49   & 92.37   \\ \hline
    LW\_COLOR & N/A       & 363.9              & 113.2              & 91.42   & 92.25   \\ \hline
    \end{tabular}
\normalsize
\label{Table:summaries}
\end{table}

Finally, we applied one implementation of the simultaneous localization and mapping technique (SLAM) to this dataset, at the same frame resolution as for the appearance-based  localization discussed in this paper. We chose the ``EKF Mono SLAM'' \citep{Civera}, which uses an Extended Kalman Filter (EKF) with 1-point RANSAC.  We chose this implementation for three reasons: a) it is a monocular SLAM technique, so comparison with the single-camera approach is fairer; b) the authors of this package report error estimates -- in the form of error distributions; and c) the errors from video with similar resolutions (240 $\times$ 320) to ours were reported as being below 2 m for some sequences \citep{Civera} in their dataset.

The results of the comparison were surprising, and somewhat unsatisfactory. The challenging ambiguity of the sequences in the RSM dataset, and possibly the low resolution of the queries, might explain the results. The feature detector, a FAST corner detector \citep{rosten_2006_machine}, produced a small number of features in its original  configuration. We lowered the feature detection threshold until the system worked on a small number of frames from each sequence. Even with more permissive thresholds, the average number of FAST features averaged only 20 across our experiments. This small number of features led to inaccuracy in the position estimates, causing many of the experimental runs to stop when no features could be matched. The small number of features per frame is also not comparable with the feature density of the methods described in this paper, where an average of 2,000 features per frame was obtained for the ``dense'' approaches. Dense SLAM algorithms might fare better.

\section{Discussion} \label{sec:discussion}
The performance comparisons shown in the cumulative error distributions of Fig.~\ref{fig:CDFAll} would seem a fairly natural means of capturing localization performance. Yet, they do not suggest large differences in terms of the AUC metric (Table~\ref{Table:summaries}), given the large diversity in the complexity of the indexing methods.  However, absolute position estimation errors tell a different story: average absolute errors are as high as 4 metres for the worst performing method (HOG3D), and just over 1.3 metres for the best performing method (SF\_GABOR), if the same camera is used.  The best performance compares very favourably with reported errors in positioning from multi-point WiFi signal strength measurements using landmark-based recognition that employs multiple (non-visual) sensing \citep{Shen}.  Indeed, it is very likely that the size of the errors we have observed can be reduced by incorporating simple motion models and a tracker, in the form of a Kalman filter.

A surprising result was that good levels of accuracy were obtained for images as small as $208\times 117$ pixels.  This suggests that relatively low-resolution cameras can be used to improve the performance of indoor localization systems.  Being able to use such low resolutions of image reduces the indexing time, storage, power and bandwidth requirements.   

\section{Conclusion \& Future Work}
 The advent of wearable and hand-held cameras makes appearance-based localization feasible. Interaction between users and their wearable device would allow for new applications such as localization, navigation and semantic descriptions of the environment. Additionally, the ability to crowdsource ``visual paths'' against which users could match their current views is a realistic scenario given ever improving connectivity.

We evaluated several types of descriptor in this retrieval-based localization scenario, achieving errors as small as 1.30 m over a 50 m distance of travel. This is surprising, given that we used low-resolution versions of our images, and particularly since our RSM dataset also contains very ambiguous indoor scenes.

We are currently working on enlarging the RSM database, by including larger numbers of journeys.  A future goal will be to!h mitigate the effects of partial occlusion between different views of the same physical location.  For example, face-detection might be applied to identify when and where people are within the scene acquired along a users' journey; we would avoid generating descriptors that covered these regions of image space. Other movable objects (chairs, trolleys) could also be actively detected and removed from indexing or queries.

The challenges associated with searching across video from multiple devices would still need to be solved. We can see from Section \ref{sec:Performance} that between-device queries have much higher error than within-device queries. This problem can be solved by either capturing and indexing data from a variety of devices for the same journeys, or by learning a mapping between devices. Another obvious strand of work would be to incorporate information from other sources, such as RSSI indicators, to reduce localization error.

Finally, we are exploring ways to combine the appearance-based technique described in this paper with SLAM and its variants. Doing this would allow geometric models from independent point-cloud sets to be associated with each other, allowing the continuous updating of the models that describe a physical space. Multiple geometric models, acquired from otherwise independent journeys, would support more detailed and reliable descriptions of an indoor, navigable space. It would also allow better interaction between the users of a building with its features, and with each other.

Our long-term goal is to convey the information acquired by sighted users to help people with visual impairment; this would require creating and updating rich descriptions of the visual and geometric structure of a physical space. This could be used in the making of indoor navigational aides, which would be rendered through haptic or audio interfaces, making the planning of journeys easier for the visually impaired.

\section*{Acknowledgements}
Jose Rivera-Rubio's research is supported by an EPSRC DTP scholarship. The dataset hosting is supported by the EPSRC V\&L Net Pump-Priming Grant 2013-1. We are also grateful to members of the Imperial College's Biologically Inspired Computer Vision group for their help and collaboration, and to Dr Riccardo Secoli and Dr Luke Dickens for their valuable input and useful discussions.


\end{document}